\documentclass[conference]{IEEEtran}
\IEEEoverridecommandlockouts
\usepackage{amsmath,amssymb,amsfonts}
\usepackage{algorithmic}
\usepackage{graphicx}
\usepackage{textcomp}
\usepackage{cite}
\usepackage{multirow}
\usepackage{booktabs}
\usepackage{tabularx}
\usepackage{xcolor}
\def\BibTeX{{\rm B\kern-.05em{\sc i\kern-.025em b}\kern-.08em
    T\kern-.1667em\lower.7ex\hbox{E}\kern-.125emX}}
\begin{document}

\title{RTFormer: Re-parameter TSBN Spiking Transformer\\
}

\author{
\IEEEauthorblockN{1\textsuperscript{st} Hongzhi Wang}
\IEEEauthorblockA{\textit{School of Software Technology} \\
\textit{Zhejiang University}\\
Ningbo, China \\
whzzju@zju.edu.cn}
\and
\IEEEauthorblockN{2\textsuperscript{nd} Xiubo Liang* \thanks{*Corresponding author}}

\IEEEauthorblockA{\textit{School of Software Technology} \\
\textit{Zhejiang University}\\
Ningbo, China \\
xiubo@zju.edu.cn}
\and
\IEEEauthorblockN{3\textsuperscript{rd} Mengjian Li}
\IEEEauthorblockA{ \textit{Research Center for Data Hub and Security}\\
\textit{Zhejiang Lab}\\
Hangzhou, China \\
limengjian@zhejianglab.com
}
\and
\IEEEauthorblockN{4\textsuperscript{th} Tao Zhang}
\IEEEauthorblockA{\textit{School of Software Technology} \\
\textit{Zhejiang University}\\
Ningbo, China \\
taozz@zju.edu.cn}

}

\maketitle

\begin{abstract}
The Spiking Neural Networks (SNNs), renowned for their bio-inspired operational mechanism and energy efficiency, mirror the human brain's neural activity. Yet, SNNs face challenges in balancing energy efficiency with the computational demands of advanced tasks. Our research introduces the RTFormer, a novel architecture that embeds Re-parameterized Temporal Sliding Batch Normalization (TSBN) within the Spiking Transformer framework. This innovation optimizes energy usage during inference while ensuring robust computational performance. The crux of RTFormer lies in its integration of re-parameterized convolutions and TSBN, achieving an equilibrium between computational prowess and energy conservation. Our experimental results highlight its effectiveness, with RTFormer achieving notable accuracy on standard datasets like ImageNet (80.54\%), CIFAR-10 (96.27\%), and CIFAR-100 (81.37\%), and excelling in neuromorphic datasets such as CIFAR10-DVS (83.6\%) and DVS128 (98.61\%). These achievements illustrate RTFormer's versatility and establish its potential in the realm of energy-efficient neural computing.
\end{abstract}


\section{Introduction}
Inspired by the human brain, deep Artificial Neural Networks (ANNs) have garnered significant success, particularly in areas such as computer vision \cite{object1,object2} and nature language processing \cite{nlp1,nlp2,nlp3,nlp4} . However, these accomplishments come at a considerable computational cost. ANNs consume approximately 12 times \cite{Attentionspiking} more energy than the human brain, rendering high-energy models tough to deploy onto resource-constrained appliances, for example, smartphones and IoT devices. Utilising the brain's efficient computational paradigm to create low-energy neural networks on such platforms holds significant value.

\textbf{Why SNN? } Spiking Neural Networks (SNNs) present themselves as paragons of energy efficiency in the computational realm. While structurally echoing the design of traditional ANNs, SNNs diverge in their unique handling of data through discrete binary events. The zero denotes a dormant state, whereas one signals the firing of a neuron, a spike that conveys information. This binary data processing leads to sparse activations within the network, ensuring that energy is expended only when necessary. Such efficiency is not merely incidental but a core feature of SNNs, enabling them to operate with a fraction of the power required by their ANN counterparts and thus answering the call for sustainable and energy-conscious computing.\par

\textbf{How can we design a structure in which inference is more energy efficient? } Addressing the question of how to craft a more energy-efficient structure for inference leads us to the inception of the Spatial-Temporal Core. The Spatial Core streamlines the convolutional process by employing structurally reparameterized convolutions, significantly reducing the computational burden during inference without compromising the integrity of learned features. Simultaneously, the Temporal Core introduces the concept of Temporal Sliding Batch Normalization (TSBN), which tailors the batch normalization process to the temporal aspects of data, ensuring that the network remains responsive to the temporal dynamics inherent in real-world scenarios. Together, these cores form a robust framework that not only excels in energy efficiency but also maintains high fidelity in data processing, making it an ideal candidate for deployment in energy-constrained environments like neuromorphic hardware.

\textbf{What's the meaning of intaking the ST-Core? }The Spatial-Temporal Core is not just a technological innovation; it's a conceptual shift towards creating neural networks that operate with a level of energy efficiency akin to the human brain. By drawing inspiration from nature's most sophisticated computing machine, we aim to bridge the gap between the computational prowess of deep learning models and the energy limitations of the devices they run on. This synergy of spatial efficiency and temporal precision paves the way for the next generation of neural network models that are both powerful and sustainable, ready for deployment in the increasingly connected and mobile world we live in.

Our contributions are summarized below:\par

\begin{itemize}
\setlength{\itemsep}{0pt}
\setlength{\parsep}{0pt}
\setlength{\parskip}{0pt}
    \item  We introduce the Spatial-Temporal Core, a harmonious fusion of structurally reparameterized convolutions (Spatial) and dynamic temporal batch normalization (Temporal) , crafted to deliver enhanced spatial efficiency and temporal adaptability, thereby elevating neuromorphic computing to new heights of processing excellence.
    \item  We introduce the TSBN, an ingenious mechanism that aligns batch normalization with the temporal dimension of data, allowing for precise, context-sensitive normalization across sequential inputs, thereby bolstering the temporal coherence and predictive performance of neural networks.
    \item Extensive experiments confirming the proposed architecture's superiority over state-of-the-art SNNs on neuromorphic and non-neuromorphic datasets, underlining its practical significance in advancing Spatial-Temporal data processing.
\end{itemize}

\section{Related Works}

\subsection{SNN Learning Methods}
Spiking Neural Networks (SNNs) are heralded as the third generation of neural network models due to their biological fidelity, intrinsic event-driven computation, and energy efficiency on neuromorphic platforms \cite{1997}. These characteristics have catalyzed a surge in SNN research, positioning them as formidable contenders to their Artificial Neural Networks (ANNs) counterparts.

The fundamental divergence between SNNs and ANNs stems from the employment of spiking neurons as the fundamental computational unit, which facilitates biological interpretability and adeptness in processing temporal information \cite{STBP,PLIF,GLIF,MLF,PSN}. ANNs, with their robust gradient backpropagation training frameworks, contrast with SNNs, which predominantly utilize two training paradigms: ANN-to-SNN conversion and direct training with surrogate gradients. The conversion approach \cite{ann2snn_bu,ann2snn_li,ann2snn_cao} entails substituting a pre-trained ANN’s ReLU layers with spiking neurons, necessitating fine-tuning of hyperparameters to retain accuracy. Nevertheless, this technique is limited by extended conversion time-steps and the architectural rigidity of the source ANN. To circumvent these limitations, \cite{STBP} employs surrogate gradients to facilitate direct SNN training, yielding high accuracy within minimal temporal intervals. Methodologies have facilitated breakthroughs across domains, with Spiking-Yolo \cite{spikingyolo} and EMS-yolo \cite{emsyolo} paving the way in object detection, Spiking-UNet \cite{spikingunet} advancing semantic segmentation, SpikingGPT \cite{spikinggpt}, and SpikingBert \cite{spikingbert} emerging in language modeling, and SpikingGAN \cite{spikinggan} introducing generative capabilities. For graph-based learning, SpikingGCN \cite{spikinggcn} and SpikingGAT \cite{spikinggat} have shown promise. The advent of neuromorphic chips such as TrueNorth \cite{True}, Loihi \cite{loihi}, and Tianjic \cite{TianJic} further underscores the potential of SNNs to become prevalent in near-term computational ecosystems.

\subsection{Transformer Architecture In SNNs}
Thanks to the well-established and effective network architectures in ANNs, SNNs can utilise them to construct high-performance models, such as \cite{spikingResnet,sewresnet,Attentionspiking,msresnet}. The attention mechanism, currently the most efficient method in ANNs, has also been integrated into SNNs, including the implementation of the Transformer, its most classic network architecture. Spikformer \cite{spikformer} is the initial directly-trained Transformer within SNNs. It adopts a new spike-form self-attention named Spiking Self Attention (SSA).
However, the current configuration of the Spikformer, which includes residual \cite{resnet} connections, still involves non-spike computation. Therefore, the Spike-Driven \cite{spike-driven} Transformer presents novel structures for preserving the spike computation. The issue of non-spike computation is resolved by Spike-Driven Transformer through introducing Spike-Driven Self-Attention (SDSA). The integration of attention mechanisms within SNNs has facilitated the adaptation of Transformer architectures to the SNN paradigm. However, previous implementations have not adequately addressed constraints encountered during inference. To this end, we propose RTFormer, which leverages a reparameterization strategy to ensure that SNNs benefit from reduced parameter complexity during inference, thereby enhancing deployment efficiency.

\subsection{BatchNormalization In SNNs}
In the realm of SNNs, the integration of Batch Normalization (BN) techniques has been pivotal in mitigating challenges associated with training dynamics, such as the vanishing or exploding gradient problem. One innovative approach, termed Batch Normalization Through Time (BNTT), was proposed by \cite{BNTT}. BNTT uniquely calculates BN statistics and parameters independently at each time-step, enhancing the network's adaptability to instantaneous changes. However, this method may not fully account for the temporal correlations present in input spike sequences, potentially overlooking crucial sequential information. To address this limitation, \cite{TDBN} introduced the threshold-dependent Batch Normalization (tdBN) methodology. This technique consolidates BN statistics and parameters across the temporal dimension, thus maintaining the conventional BN's benefits while accommodating the temporal structure inherent in SNNs. By aggregating data temporally, tdBN circumvents the instability often encountered in gradients during SNN training. Further expanding upon these developments, the Temporal Effects Batch Normalization (TEBN) method, conceived by \cite{TEBN}, has refined the approach by merging data along the temporal axis for shared BN statistics. TEBN then introduces temporal dynamics into the BN process by applying distinct scaling weights. This method captures the essential temporal dynamics, thereby providing a more nuanced normalization process that aligns with the temporal nature of SNNs. We introdece the TSBN to selectively leverage the accumulated pre-synaptic inputs in the temporal domain, consistent with the properties of LIF neurons.

\section{Method}
We introduce the RTFormer, a novel fusion of the Transformer architecture with the Re-parameter  and Temporal Sliding Batch Normalization(TSBN). This section will begin by providing a concise overview of spike neurons' working principles, followed by an in-depth exploration of the Spatial-Temporal Core and the Spiking Guided Attention (SGA) module. Finally, we will discuss the energy consumption aspect.
\subsection{Preliminaries}

In SNNs, spike neurons control the release of spikes based on a threshold. In this paper we use LIF \cite{LIF} neurons, which work in the following way:
\begin{align}
   &U[t] = V[t-1] + \frac{1}{k_{\tau}}\left(X[t] - (V[t-1]-V_{reset})\right) \\
&S[t] = \mathcal{H}(U[t]-V_{th}) \\
&V[t]=U[t]~(1-S[t]) + V_{reset}S[t] 
\end{align}
 where $k_{\tau}$, $V_{th}$, and $V_{reset}$ represent the decay factor, firing threshold, and reset membrane potential, respectively, which are pre-set to default values. The notation $X[t]$ refers to the input at time step $t$, while $U[t]$ denotes the membrane potential. The function $\mathcal{H}$($\cdot$) represents the Heaviside step function. The spike output, denoted by $S[t]$, is calculated based on the membrane potential and the threshold. Additionally, $V[t]$ and $V[t-1]$ signify the temporal output at time t.

In our study, we employ the spikingjelly \cite{fang2023spikingjelly}, utilizing default values for $k_{\tau}$, $V_{th}$, and  $V_{reset}$, specifically set to 2.0, 1.0, and 0, respectively.

\begin{figure}

 \centering
     \setlength{\abovecaptionskip}{0pt}%    
    \setlength{\belowcaptionskip}{10pt}%

 \label{fig:STCore}
 \includegraphics[width=75mm]{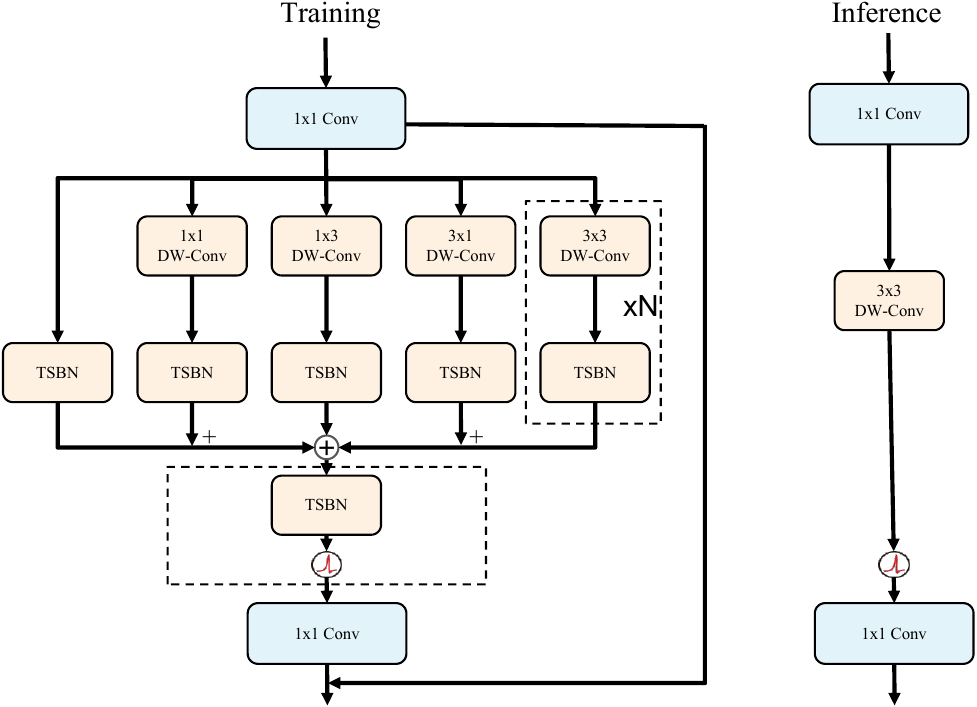}
 \caption{Illustration of the structural reparameterization and simplification from a complex multi-branch system to a streamlined model. The top dashed box represents the parameters of the TSBN incorporated into Conv3, and the bottom dashed box represents the parameters of the TSBN incorporated into the trainable threshold.}
\end{figure}

\subsection{Spatial-Temporal Core}
In the exploration of SNNs, our innovative design, termed the "Spatial-Temporal Core," shown in Fig.\ref{fig:STCore}, addresses the nuanced realms of spatial and temporal processing. This sophisticated framework divides the architecture into two synergistic components: the "Spatial Core" and the "Temporal Core." Each core is uniquely adapted to manage distinct aspects of data processing - the former concentrating on spatial features and the latter on temporal dynamics.\par

\textbf{Spatial Core. }The "Spatial Core" echoes the principles of structural reparameterization, akin to approaches seen in Artificial Neural Networks (ANNs). Here, we utilize depthwise convolutions (DW-Conv) that are strategically reparameterized to reduce the complexity of the model during inference. This innovative arrangement entails parallel DW-Conv layers with diverse kernel sizes, notably 1x1 and 3x3, enhancing spatial feature extraction efficiency. The transformation from four consecutive 3x3 layers to a trio of "Spatial Core" units signifies a leap in spatial detail capture, all while maintaining a lean model structure.\par

In the innovative architecture depicted in Fig.\ref{fig:STCore}, the STCore is structured with five concurrent branches, each contributing unique convolutional parameters to the network's composite function. Notably, one of these branches—the identity branch—functions effectively as a 1 × 1 convolution, utilizing an identity matrix as its kernel. This integration ensures that each branch's convolutional characteristics are distinctly represented. The fusion of these diverse convolutional influences is succinctly captured in Eq.\ref{eq:branches}:

\begin{equation}
y = \sum_{i=1}^n TSBN_i(x*W^{(i)},\mu_i,\sigma_i,\gamma_i,\beta_i)
\label{eq:branches}
\end{equation}

Here, the index $i$ extends over $n$ distinct branches, with this work specifically focusing on $n = 5$. The variable $x$ symbolizes the input, while $y$ represents the resultant output. The operator  $"*"$ denotes the convolution operation, and $(i)$ signifies the convolutional kernelassociated with the $i_{th}$ branch. The parameters $\mu, \sigma, \gamma, \beta$ correspond to the accumulated mean, standard deviation, as well as the learned scaling factor and bias, respectively, derived from the BN layer that follows each convolutional operation. These parameters are crucial for the TSBN process, which refines the data at each branch, ensuring a harmonized and effective integration of the temporal dynamics into the network's overall learning process.
 
\textbf{Temporal Core. }Conversely, the "Temporal Core" integrates Temporal Sliding Batch Normalization (TSBN) with the adjustable thresholds of spiking neurons. This integration is a pivotal aspect of our approach, aligning with the innate dynamics of temporal information processing inherent in SNNs. By incorporating the TSBN parameters directly into the neurons' thresholding mechanism (denoted as $V_{th}$), our model gains a robustness in handling temporal sequences, which is essential for cognitive functions.\par
In practice, a sliding window mechanism judiciously controls the extent of Batch Normalization across time, allowing for refined data processing at each stage. Diverging from traditional BN methods like tdBN and BNTT, our focus is on data closer to the current timestep, ensuring a more context-sensitive normalization approach. This methodology enhances the network's capability to adeptly respond to dynamic temporal shifts, as captured in Eq.\ref{eq:tsbn}.\par 

\begin{equation}
    y_{[t-w:t]} =  \gamma(t)\frac{x_{[t-w:t]}-\mu_{[t-w:t]}}{\sqrt{\sigma_{[t-w:t]}^{2}+\epsilon}}+\beta(t)
    \label{eq:tsbn}
\end{equation}

Here, $x_{[t-w:t]}$ and $y_{[t-w:t]}$ signify the input and output, respectively, spanning a temporal window of width $w$. The variable $t$ specifies the current timestep, anchoring the window's position within the sequence. $\gamma(t)$ and $\beta(t)$ serve as the scale and shift factors at each specific timestep $t$. $\mu_{[t-w:t]}$ and $\sigma_{[t-w:t]}$ are the statistical core of this equation, representing the mean and variance computed over the inputs within the sliding window from $t-w$ to $t$, $\epsilon$ is a small number to avoid dividing zero .

Moreover, we fold the TSBN into the neuron's threshold 
 , enabling an efficient integration of temporal normalization into the spiking mechanism of the neurons. This folding process is encapsulated in Eq.\ref{eq:vth}, transforming $V_{th}$ into a trainable parameter , which dynamically adapts to the temporal normalization.
\begin{equation}
    {s}^{(t)} = 
    \begin{cases}
        1 & \text{if } \gamma(t)\frac{x_{[t-w:t]}-\mu_{[t-w:t]}}{\sqrt{\sigma_{[t-w:t]}^{2}+\epsilon}}+\beta(t) > V_{\rm th} \\
        0 & \text{otherwise}
    \end{cases}. 
    \label{eq:compare}
\end{equation}

\begin{equation}
    {\tilde{V}}_{\rm th} = \frac{(V_{\rm th} - {\beta}(t)){\sqrt{{\sigma}_{[t-w:t]}^2}}}{{\gamma}(t)} + {\mu}_{[t-w:t]}
    \label{eq:vth}
\end{equation}

Here, $V_{th}$ is the threshold of spiking neuron, ${\tilde{V}}_{\rm th}$  is its trainable counterpart in this work. $s^{(t)}$ is the spiking output matrix at timestep t.

The "Spatial-Temporal Core" stands as a hallmark of innovation in SNN architecture. Its ability to fluidly navigate both spatial and temporal dimensions positions it as a vital development towards emulating the sophisticated computational capabilities of the human brain. This structure not only ensures model efficiency but also accentuates the biological resemblance and pulse-like nature of SNNs, underscoring its potential for real-world applications where both performance and biologically-inspired functionality are paramount.

\begin{table*}[htbp]
  \centering
  \caption{Experiments on ImageNet. 'T' denotes the timestep. The architecture abbreviations 'S-V,' 'S-R,' and 'S-T' correspond to Spiking VGG, Spiking Resnet, and Spiking Transformer, respectively.}
    \begin{tabular}{clllrrrr}
    \toprule
    \multicolumn{1}{l}{Dataset} & Methods & Type  & Architecture & \multicolumn{1}{l}{T} & \multicolumn{1}{l}{Param(M)} & \multicolumn{1}{l}{Acc(\%)} & \multicolumn{1}{l}{Energy(mJ)} \\
    \midrule
    \multirow{11}[8]{*}{ImageNet} & RMP \cite{RMP}   & ANN2SNN & S-V-16 & 4096  & 138.4 & 73.09 & 49.86 \\
          & Calibration \cite{Cali} & ANN2SNN & S-V-16 & 2048  & 138.4 & 75.32 & 25.98 \\
          & SEW-ResNet \cite{sewresnet} & SNN training & S-R-152 & 4     & 60.19 & 69.26 & 12.89 \\
          & MS-ResNet \cite{msresnet} & SNN training & S-R-104 & 4     & 77.28 & 76.02 & 10.19 \\
          & Att-MS-ResNet \cite{Attentionspiking} & SNN training & S-R-104 & 4     & 78.37 & 77.08 & 7.3 \\
\cmidrule{2-8}          & tdBN  \cite{TDBN} & SNN training & S-R-34 & 6     & 21.79 & 63.72 & 6.39 \\
          & TEBN  \cite{TEBN} & SNN training & S-R-34 & 4     & 21.79 & 64.29 & 7.05 \\
          & MPBN  \cite{MPBN} & SNN training & S-R-34 & 4     & 21.79 & 64.71 & 6.56 \\
\cmidrule{2-8}          & spikformer \cite{spikformer} & SNN training & S-T-8-768 & 4     & 66.34 & 74.81 & 21.47 \\
          & spike-driven \cite{spike-driven} & SNN training & S-T-8-768* & 4     & 66.34 & 77.07 & 6.09 \\
\cmidrule{2-8}          & This work & SNN training & S-T-8-768 & 4     & 58.86 & \textbf{80.54} & 5.59 \\
    \bottomrule
    \end{tabular}%
  \label{tab:imagenet}%
\end{table*}%

\subsection{Analyse Of Energy Consumption}

In ANNs, computational demands largely stem from Floating-point Operations (FLOPs), predominantly due to Multiply and Accumulate (MAC) operations. SNNs, however, primarily rely on Accumulate (AC) operations, which reduces the need for MAC operations. This shift not only cuts down on FLOPs but also aligns with SNNs' energy-efficient ethos by reducing power usage.

Yet, MAC operations remain a factor in the initial stages of data processing, where raw images are converted into spike-encoded formats. To gauge energy use, an assessment of MAC and AC operations throughout the network's computational processes is essential.

The total energy consumption ($E$) can be expressed as follows:

\begin{align}
E = E_{MAC} \times FL_{conv} + E_{AC} \times \\ \nonumber  (FL_{STCore} + FL_{SGA} + FL_{STMLP})
\end{align}

Here, $E_{MAC}$ and $E_{AC}$ represent the energy consumption associated with MAC and AC operations, respectively. Experimental measurements have determined $E_{MAC}$ to be approximately 4.6 picojoules (pJ), and $E_{AC}$ to be approximately 0.9 picojoules (pJ), based on testing conducted on 45nm technology \cite{45nm}, $FL_{\cdot}$ denotes the float number of the specific layer.

Calculating energy consumption offers an accurate measure of the computational and energy efficiency enhancements delivered by our framework, considering the interplay of both MAC and AC operations across the processing pipeline.

\section{Experiments}
\begin{figure}[h]

 \centering
      \setlength{\abovecaptionskip}{0pt}%    
    \setlength{\belowcaptionskip}{0pt}%

 \label{fig:attention}
 \includegraphics[width=75mm]{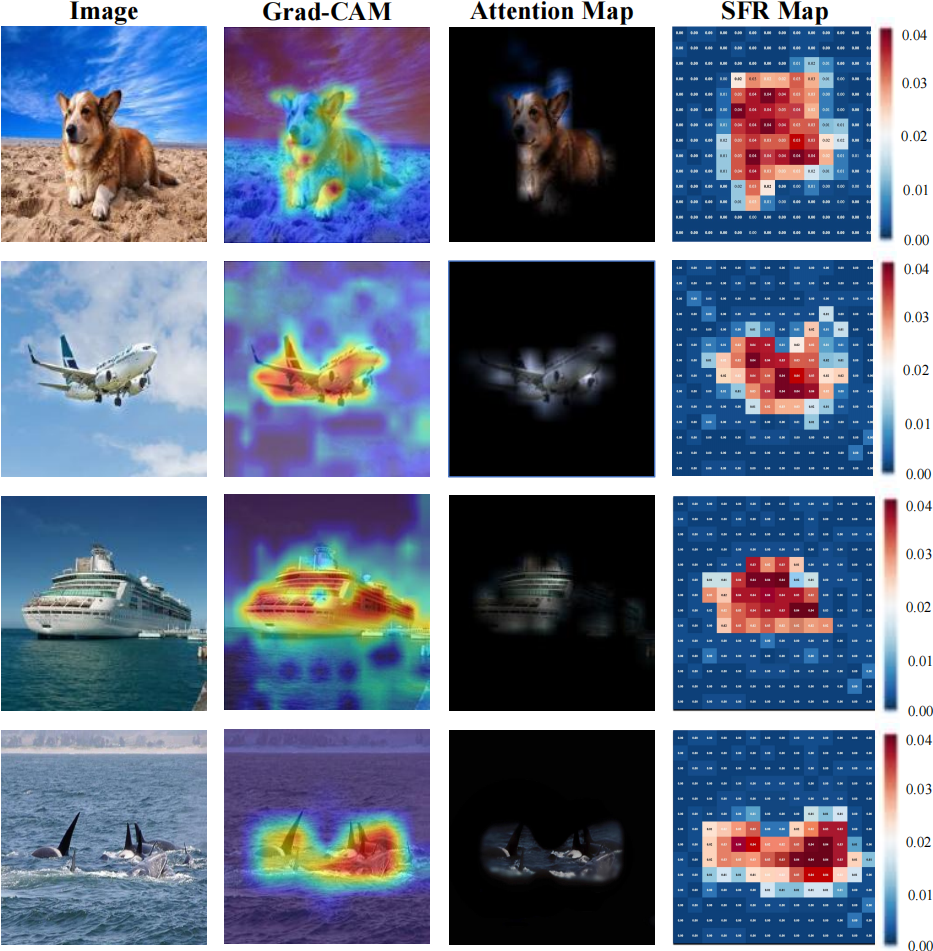}
  \caption{
Display of a comparative visualization across four columns for a series of images. Visualization comprises original images, Grad-CAM representations, Attention Maps , and Spiking Fire Rate (SFR) maps.}
\end{figure}

Our experimental evaluation encompasses both non-neuromorphic datasets like CIFAR10, CIFAR100, and ImageNet, as well as neuromorphic datasets such as CIFAR10-DVS and DVS128 Gesture. The visualizations of these results are presented in Fig.\ref{fig:attention}, with ImageNet findings detailed in Tab.\ref{tab:imagenet}. Results for other datasets are compiled in Tab.\ref{tab:datasets}, while the outcomes of our ablation studies are summarized in Tab.\ref{tab:ablation}. Additionally, visual representations of these ablation studies can be found in Fig.\ref{fig:plot}. 

\subsection{Non-neuromorphic Datasets Classification}
\subsubsection{ImageNet} \par
\textbf{Dataset Description. } The ImageNet dataset, a cornerstone in the field of computer vision, consists of approximately 1.3 million images spanning 1,000 classes for training, alongside 50,000 validation images. \par

\textbf{RTFormer's performance. } As shown in Table \ref{tab:imagenet}, the RTFormer model with TSBN and structurally reparameterized DW-conv achieves remarkable accuracy. Specifically, the Spikformer-8-768 model, equipped with 58.86M parameters, attains a top-1 accuracy of 80.54\%, which is a notable enhancement over previous SNN models like SEW-ResNet and MS-ResNet. This leap in performance is also accompanied by a reduction in energy consumption, underlining the efficiency of the model's transformer architecture and optimized components. \par
\textbf{Comparision with BN methods in SNNs. }  RTFormer significantly surpasses the tdBN method, achieving a higher accuracy rate of 80.54\% compared to 64.29\% obtained by tdBN. This stark difference in performance indicates the effectiveness of the architectural and BN improvements. Moreover, the energy consumption of "This work" is lower (5.59mJ) compared to tdBN (7.05mJ), highlighting improved efficiency. Similarly, RTFormer showcases superior accuracy over TEBN, with a 13.87\% increase in top-1 accuracy. RTFormer achieves an accuracy rate that is 13.45\% higher than that of the MPBN method. The energy savings are also notable, with RTFormer consuming less energy, thereby presenting a strong case for the enhancements brought by the TSBN and Re-parameter Transformer architecture. \par
\textbf{Comparision with Transformers in SNN. } The Spikformer shows competitive performance, achieving a 74.81\% accuracy rate. However,  RTFormer outperforms Spikformer by 3.35\%, which is a significant margin in the realm of deep learning models. The Spike-Driven Transformer model, another variant of the Spiking Transformer, achieves a 77.07\% accuracy rate, which is commendable. Nonetheless, RTFormer has a slight edge with an accuracy of 78.16\%. The energy efficiency of our architecture is notably better, with an energy consumption of 5.59mJ compared to the spike-driven model's 6.09mJ, showcasing that the model's improvements do not come at the cost of increased energy usage.

\begin{table*}[tbp]
  \centering
  \caption{Experiments on both Static and DVS datasets. In this table, 'Ts' refers to the Timestep for Static datasets, while 'Td' indicates the Timestep for Neuromorphic (DVS) datasets. The architecture abbreviations 'S-V,' 'S-R,' and 'S-T' correspond to Spiking VGG, Spiking Resnet, and Spiking Transformer, respectively. Asterisked results (*) represent outcomes from our implementations of these methods. Detailed hyperparameters used in these experiments are meticulously documented in the appendix for reference.}
    \begin{tabular}{lllllllll}
    \toprule
    \multicolumn{1}{c}{\multirow{2}[4]{*}{Methods}} & \multicolumn{1}{c}{\multirow{2}[4]{*}{Architecture}} & \multicolumn{1}{c}{\multirow{2}[4]{*}{Param(M)}} & \multicolumn{1}{c}{\multirow{2}[4]{*}{Ts}} & \multicolumn{2}{c}{Acc} & \multicolumn{1}{c}{\multirow{2}[4]{*}{Td}} & \multicolumn{2}{c}{Acc} \\
\cmidrule{5-6}\cmidrule{8-9}          &       &       &       & \multicolumn{1}{c}{cifar10} & \multicolumn{1}{c}{cifar100} &       & \multicolumn{1}{c}{cifar10-dvs} & \multicolumn{1}{c}{dvs 128} \\
    \midrule
    RMP \cite{RMP}  & S-V-16 & 138.4 & 4096  & 93.63 & 70.93 &  -    &  -    &  - \\
    Calibration \cite{Cali} & S-V-16 & 138.4 & 2048  & 95.79 & 77.87 &  -    &  -    &  - \\
    SEW-ResNet \cite{sewresnet} & S-R-21 & 21.79 & 4     & 95.34* & 78.32* & 16    & 74.4  & 97.9 \\
    MS-ResNet \cite{msresnet} & S-R-18 & 11.69 & 4     & 94.79* & 78.15* & 16    & 75.56 & 97.54* \\
    Att-MS-ResNet \cite{Attentionspiking} & S-R-18 & 11.87 & 4     & 95.07* & 77.89* & 20    & 77.35* & 98.23 \\
    \midrule
    tdBN \cite{TDBN}  & S-R-19 & 12.63 & 4     & 92.92 & 70.86 & 16    & 67.8  & 96.9 \\
    TEBN \cite{TEBN}  & S-R-19 & 12.63 & 4     & 94.7  & 76.13 & 10    & 83.3* & 97.95* \\
    MPBN \cite{MPBN}  & S-R-19 & 12.63 & 2     & 96.05* & 79.51 & 10    & 74.4  & 98.26* \\
    \midrule
    Spikformer \cite{spikformer} & S-T-4-384 & 9.32  & 4     & 95.19 & 77.86 & 16    & 80.9  & 98.3 \\
    Spike-Driven \cite{spike-driven} & S-T-4-384 & 9.32  & 4     & 95.6  & 78.4  & 16    & 80    & 97.9* \\
    \midrule
    \multirow{2}[2]{*}{This work} & \multirow{2}[2]{*}{S-T-4-384} & \multirow{2}[2]{*}{7.93} & 4     & 96.27 & 81.37 & 16    & 83.6  & 98.61 \\
          &       &       & 2     & 96.12 & 80.9  & 10    & 82.9  & 98.61 \\
    \bottomrule
    \end{tabular}%
  \label{tab:datasets}%
\end{table*}%

\subsubsection{CIFAR} 
\textbf{Dataset Description. } The CIFAR-10 dataset is a well-known collection of 60,000 32x32 color images split into 10 classes, with each class represented by 6,000 images. The CIFAR-100 dataset, while similar in structure to CIFAR-10, offers a more challenging task with its 100 classes, each comprising 600 images, also totaling 60,000.  Both datasets, developed by the Canadian Institute for Advanced Research, serve as fundamental benchmarks for machine learning and computer vision, facilitating the development and validation of innovative image classification models.\par

\textbf{Comparision with previous works. } Firstly, as shown in Tab.\ref{tab:datasets} when benchmarked against previous SNN models like RMP and others, RTFormer demonstrates superior performance. While traditional SNNs like RMP and Calibration have laid the groundwork in the field, RTFormer capitalizes on their foundation and pushes the boundaries further. For instance, on CIFAR-100, RTFormer achieves an accuracy of 81.37\%, which is a substantial improvement over the 70.93\% and 77.87\% accuracy rates achieved by RMP and Calibration, respectively. This leap in performance can be attributed to RTFormer's more sophisticated temporal dynamics capturing capabilities and its optimized training methodologies.\par

\textbf{Comparision with BN methods in SNN. } Secondly, in comparison with other SNN models employing various BN techniques, RTFormer stands out for its effective use of TSBN. This technique provides RTFormer with an edge, allowing for better normalization of the neuron's output across different timesteps, which is crucial for datasets with a high degree of intra-class variability like CIFAR-100. The improvement in normalization contributes to more stable and faster convergence during training, as evidenced by the higher accuracy rates when compared to the tdBN, TEBN, and MPBN methods.\par

\textbf{Comparision with Transformers in SNN. } Thirdly, against the backdrop of Spike Transformer architectures, RTFormer's refined approach shines through. RTFormer's innovative BN approach, coupled with structurally reparameterized depthwise convolutions (DWconv), significantly boosts its performance. While Spikformer and spike-driven models have shown the viability of Transformer architectures in SNNs, RTFormer optimizes these designs, achieving an impressive 96.27\% on CIFAR-10 and 81.37\% on CIFAR-100. This represents not only an improvement over the aforementioned models but also highlights the RTFormer's architectural benefits, which are particularly advantageous for complex and nuanced datasets like CIFAR-100.\par

In conclusion, RTFormer, with its strategic modifications and enhancements, stands as a testament to the potential of SNNs, particularly in processing complex visual data, and sets a new benchmark for accuracy and efficiency in the field.

\begin{figure}[h]

 \centering
     \setlength{\abovecaptionskip}{10pt}%    
    \setlength{\belowcaptionskip}{0pt}%

 \label{fig:plot}
 \includegraphics[width=85mm]{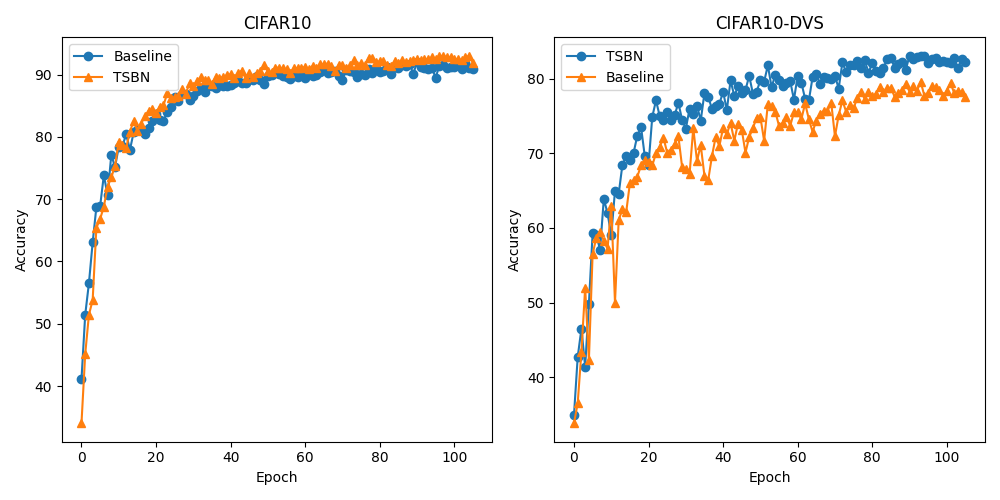}
  \caption{The figure presents two line graphs, where the blue line represents the baseline model, and the green line indicates the performance after incorporating  TSBN. The graph on the left illustrates the results obtained on the CIFAR-10 dataset, while the right graph showcases the outcomes on the CIFAR10-DVS dataset. }
\end{figure}

\subsection{Neuromorphic Datasets Classification}
\textbf{Dataset Description. }The CIFAR10-DVS dataset is a neuromorphic version of the well-known CIFAR-10 dataset, converted using a Dynamic Vision Sensor (DVS). It presents everyday objects in a format compatible with neuromorphic vision systems, capturing temporal changes in pixel intensity. DVS128 Gesture, on the other hand, is a gesture recognition dataset specifically designed for neuromorphic processing. It comprises hand gesture data from 29 individuals under various lighting conditions, captured through a DVS camera, making it ideal for developing and testing gesture recognition models on SNNs and neuromorphic hardware.

As shown in  Tab.\ref{tab:datasets}, RTFormer exhibits unique advantages in Neuromorphic datasets, such as CIFAR10-DVS and DVS128 Gesture, which capitalize on the intrinsic features of Spiking Neural Networks (SNNs) and the architectural innovations specific to RTFormer. \par

\begin{table}[h]
  \centering
  \caption{Ablation Experiment for TSBN. In the "Architecture" column of the table, the abbreviation "S" stands for "Spiking," "R" represents "ResNet," and "T" denotes "Transformer." }
    \begin{tabular}{cccc}
    \toprule
    Dataset & Method & Architecture & Acc.(\%) \\
    \midrule
    \multirow{4}[4]{*}{CIFAR10} & Baseline & S-R-19 & 95.28\% \\
          & w/ TSBN & S-R-19 & 96.04\% \\
\cmidrule{2-4}          & Baseline\cite{spike-driven} & S-T-4-384 & 95.60\% \\
          & w/ TSBN & S-T-4-384 & 96.27\% \\
    \midrule
    \multirow{4}[4]{*}{CIFAR100} & Baseline & S-R-19 & 74.52\% \\
          & w/ TSBN & S-R-19 & 79.37\% \\
\cmidrule{2-4}          & Baseline\cite{spikformer} & S-T-4-384 & 80.90\% \\
          & w/ TSBN & S-T-4-384 & 83.60\% \\
    \bottomrule
    \end{tabular}%
  \label{tab:ablation}%
\end{table}%
\textbf{Compared to Previous Research. }RTFormer outshines prior studies, notably in neuromorphic datasets like CIFAR10-DVS and DVS128 Gesture. It demonstrates superior accuracy, a clear advancement over earlier methods like RMP and Calibration, which do not present results for these DVS datasets.

\textbf{Against Other BN Methods in SNNs. }When compared to other batch normalization techniques such as tdBN, TEBN, and MPBN, RTFormer exhibits noteworthy improvements in accuracy on neuromorphic datasets. This suggests that its approach to integrating batch normalization is more effective for handling the dynamic nature of these datasets.

\textbf{Versus Other Spiking Transformer Architectures. }RTFormer also excels in comparison to other spiking transformer architectures like Spikformer and Spike-Driven. It achieves higher accuracy rates, highlighting its effectiveness in processing the temporally rich data characteristic of neuromorphic datasets, thereby underscoring its advanced capability in handling spatiotemporal data complexities.

\subsection{Ablation Study}
To verify the effectiveness of the TSBN, a lot of ablative studies using different architecture  were conducted on the CIFAR10 and CIFAR10-DVS datasets. Table \ref{tab:ablation} clearly demonstrates that the integration of Temporal Sliding Batch Normalization (TSBN) leads to an enhancement in accuracy, regardless of whether the underlying backbone is a Spiking ResNet or a Spiking Transformer. This improvement is consistent across both Static and Neuromorphic datasets, a fact that is also visually evident in the accompanying Figure \ref{fig:plot}.

\subsection{Performance Insights of RTFormer }
\textbf{Architectural Synergy with Neuromorphic Data. } Neuromorphic datasets inherently contain temporal information that traditional static datasets lack, and RTFormer is adept at leveraging this. The model’s architecture, influenced by the Transformer design, is inherently suited for handling sequences, making it exceptionally well-aligned with the time-sensitive data in neuromorphic datasets. The RTFormer uses Spiking Transformer blocks tailored to process the spatio-temporal dynamics present in such data, enabling it to capture the nuanced temporal patterns that are pivotal for recognition tasks in neuromorphic vision. \par
\textbf{Effectiveness on Static Data. }The RTFormer's effectiveness in processing static datasets can be attributed to its novel integration of re-parameterized convolutions. The spatial core's re-parameterized convolutions adeptly capture complex spatial patterns in static data, while the TSBN, even in a non-temporal context, provides adaptive normalization that enhances the network's ability to generalize from training to unseen data. This combination not only boosts computational efficiency but also ensures a high level of accuracy, making RTFormer a versatile tool in both dynamic and static data environments. \par
\textbf{Efficient Temporal Encoding. } The RTFormer's use of Temporal Sliding Batch Normalization (TSBN) is particularly beneficial for neuromorphic datasets. This specialized BN method ensures that RTFormer's neurons maintain an optimal firing rate, preventing both the vanishing and exploding gradient problems that are common in SNNs. This allows the RTFormer to efficiently encode temporal information, a critical aspect when dealing with datasets like CIFAR10-DVS, where each pixel's intensity changes over time are encoded into spike trains.\par

\textbf{Robust to Diverse Visual Stimuli. }RTFormer's robustness to diverse visual stimuli, stemming from its Transformer roots, is evident in its neuromorphic dataset performance. The attention mechanisms allow the model to focus on the most salient features within the spike trains, enhancing its ability to discern between different gestures and visual patterns with high accuracy.

\section{Conlcusion }
 To ensure efficient inference on spiking chips, all convolutional operations within our model have been reparameterized structurally. In conjunction, we have refined the subsequent Batch Normalization (BN) technique to align with the characteristics of Leaky Integrate-and-Fire (LIF) neurons, resulting in the introduction of Temporal Sliding Batch Normalization (TSBN). By embedding TSBN into the  transformer architecture, we have crafted the RTFormer, which achieves unprecedented results on both static and neuromorphic datasets, setting new benchmarks in performance. \par

\section*{Acknowledgment}
This work was partly supported by "Pioneer" and "Leading Goose" R\&D Program of Zhejiang (2023C01045) and Ningbo Leading Talents Training Project.
\bibliography{conference}
\bibliographystyle{IEEEtran}

\vspace{12pt}

\end{document}